\documentclass[letterpaper, 10 pt, conference]{ieeeconf}

\IEEEoverridecommandlockouts                              

\let\labelindent\relax
\usepackage{enumitem}

\usepackage{array}
\usepackage{graphicx}
\usepackage{cite}
\usepackage{amsmath}
\usepackage{amssymb}
\usepackage{stmaryrd}
\usepackage{multirow}
\usepackage{algorithm}
\usepackage{algorithmic}
\usepackage[T1]{fontenc}
\usepackage{multicol}
\usepackage{stfloats}
\usepackage[caption = false]{subfig}
\usepackage{xcolor}

\usepackage{mathtools}

\usepackage{makecell}
\usepackage{scalerel}
\usepackage{array}
\usepackage{multirow}
\usepackage{dsfont}
\usepackage{url}

\newcolumntype{M}{>{\centering\arraybackslash}m{.2\textwidth}}
\newcolumntype{C}[1]{>{\centering\let\newline\\\arraybackslash\hspace{0pt}}p{#1}}
\newcolumntype{R}[1]{>{\raggedleft\let\newline\\\arraybackslash\hspace{0pt}}p{#1}}
\newcolumntype{L}[1]{>{\raggedright\let\newline\\\arraybackslash\hspace{0pt}}p{#1}}

\newcommand\Tstrut{\rule{-3pt}{2.6ex}}       
\newcommand\Bstrut{\rule[-0.9ex]{-3pt}{0pt}} 
\newcommand{\TBstrut}{\rule{-3pt}{2.6ex} \rule[-0.9ex]{-2pt}{0pt}}  

\newcommand\mydots{\hbox to 1em{.\hss.\hss.}}

\begin{document}

\title{\LARGE \bf
Self-Supervised Learning of Lidar Segmentation for Autonomous Indoor Navigation
}

\author{
Hugues Thomas$^{1}$
\quad
Ben Agro$^{1}$
\quad
Mona Gridseth$^{1}$
\quad 
Jian Zhang$^{2}$
\quad
Timothy D. Barfoot$^{1}$
\thanks{$^{1}$University of Toronto Institute
for Aerospace Studies (UTIAS), 4925 Dufferin St, Ontario,
Canada. $^{2}$Apple Inc., 
{\tt\small \{hugues.thomas, ben.agro, mona.gridseth\}@robotics.utias.utoronto.ca, jianz@apple.com, tim.barfoot@utoronto.ca}}
}

\maketitle


\IEEEpeerreviewmaketitle

\begin{abstract}

We present a self-supervised learning approach for the semantic segmentation of lidar frames. Our method is used to train a deep point cloud segmentation architecture without any human annotation. The annotation process is automated with the combination of simultaneous localization and mapping (SLAM) and ray-tracing algorithms. By performing multiple navigation sessions in the same environment, we are able to identify permanent structures, such as walls, and disentangle short-term and long-term movable objects, such as people and tables, respectively. New sessions can then be performed using a network trained to predict these semantic labels. We demonstrate the ability of our approach to improve itself over time, from one session to the next. With semantically filtered point clouds, our robot can navigate through more complex scenarios, which, when added to the training pool, help to improve our network predictions. We provide insights into our network predictions and show that our approach can also improve the performances of common localization techniques.

\end{abstract}


\section{Introduction}

In a high-traffic area or crowded building, basic navigation tasks such as localization and path planning can be difficult. To make navigation more robust to such complex environments, one can use semantic information to filter the visual cues used by each particular algorithm. For example, localization algorithms should focus on permanent objects such as walls, and path planning algorithms should not treat moving objects as if they were static as they are likely to get out of the robot's way on their own. Supervised Deep Learning approaches are good candidates for the prediction of semantic information in real-time. However, such supervised methods need to be trained on a huge number of hand-annotated examples, which are not always available.

\begin{figure}[t]
    \centering
    \includegraphics[width=0.98\columnwidth, keepaspectratio=true]{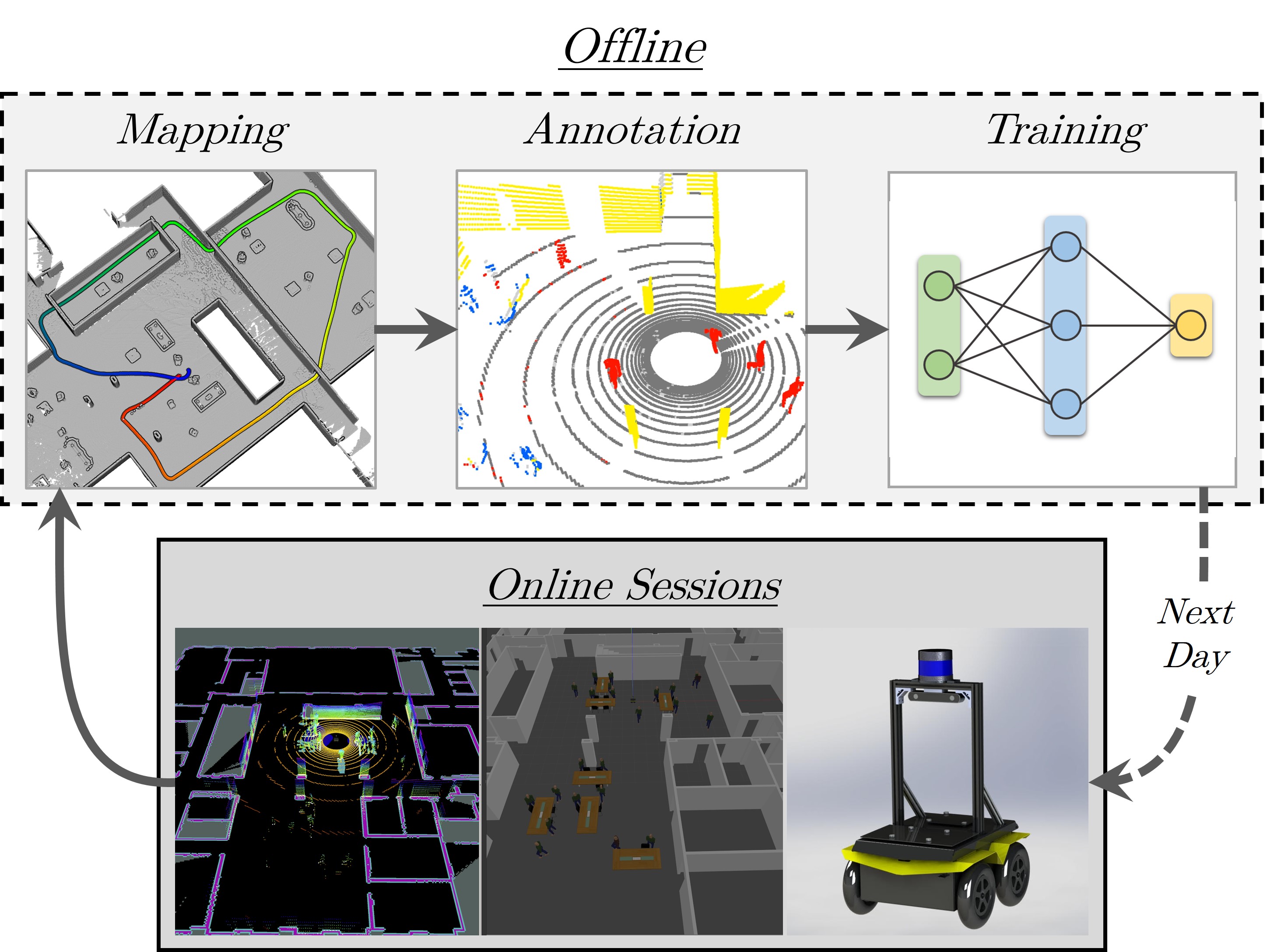}
    \caption{Illustration of our self-improving approach in a multi-session setup. First, navigation sessions are performed by a robot. After they are completed, we map and annotate them. Then we train a deep network to predict the labels. Using semantically filtered point clouds, the robot can navigate in more complex new scenarios, which will then be added to the pool of training sessions.}
    \label{fig_intro}
    \vspace{-4ex}
\end{figure}

In this paper, we propose a self-supervised learning approach for the segmentation of lidar point clouds in a long-term multi-session setup (Figure \ref{fig_intro}). Our method revolves around an automated annotation process, which labels 3D point clouds from previous sessions. A point cloud convolutional neural network can then be trained to predict semantic information in future sessions. We demonstrate the ability of our method to improve from session to session. The network predictions allow our robot to evolve in increasingly complex scenarios, and the more complex scenarios it sees, the better its predictions become.

Our annotation methodology uses two main building blocks: An ICP-based 3D SLAM method we call PointMap, able to build or localize on point cloud maps, and an algorithm estimating the occupancy probabilities in a point cloud map using ray-tracing, that we name PointRay. For each session, we annotate a map of the environment, and these annotations are projected back to the frames that were used to build the map. We are able to identify four semantic labels: \textit{ground}, \textit{permanent} (points from objects that never move like walls), \textit{shortT} (short-term movables such as people), \textit{longT} (long-term movables that are still but can be relocated between sessions, such as furniture). Some points remain \textit{uncertain} and are ignored during training. We then train a point convolution network built with KPConv \cite{thomas2019kpconv} to predict the labels given single lidar frames as input.

During online sessions, the predictions are used as a triaging system for the point clouds. We only give the \textit{permanent} and \textit{ground} points to the localization algorithm. As for the local and global planning, we simply remove the short-term movable points with the assumption that they will get out of the way. This is a first idea to demonstrate the ability of our approach, but in the future, we plan to use a more elaborate planning algorithm that incorporates both static and moving points and treats them differently.

We conduct all our experiments in a Gazebo simulation using a reproduction of the intended environment (bottom pictures in Figure \ref{fig_intro}). The simulation has the advantage of providing groundtruth for both point labeling and localization, allowing a thorough verification of our approach\footnote{The experiments were originally supposed to be conducted on a real robot platform in a busy university building. However, due to the pandemic situation, it is currently impossible to experiment with crowded scenarios in the real world.}.

After a literature review in Section \ref{sec_Related}, we describe all the components of our approach in Section \ref{sec_Method}. Then, we present our experimental results in Section \ref{sec_Exp}. Finally, we discuss concluding remarks and future work in Section \ref{sec_Conclusion}.


\section{Related Work}
\label{sec_Related}

Self-supervised learning is a form of unsupervised learning where the data provides supervision. In robotics, this term usually refers to methods using an automated acquisition of training data \cite{sofman2006improving, lookingbill2007reverse, hadsell2009learning, brooks2012self, ridge2015self, nava2019learning}. These approaches often exploit multiple sensors during the robot's operation. In our case, we only use a 3D lidar and an algorithm that provides automated annotation for the data. Deep-learning-based semantic SLAM algorithms have been proposed, using either camera images \cite{zhang2018semantic, wang2019unified}, lidar depth images \cite{chen2019suma++}, or lidar point clouds \cite{sun2018recurrent}, but they always rely on human-annotated datasets, whereas our method learns on its own.

Dewan et al. \cite{dewan2017deep} proposed to use a deep network to help detect movable points. However, they chose a 2D architecture using lidar depth images, FastNet, and only predict an objectness score from a human-annotated training set. They infer the movable probability from the motion between two consecutive frames and thus need to calculate odometry first. Our predictions come from a single frame, and therefore, they can be used to improve odometry. Furthermore, we make a distinction between short-term and long-term movables, similarly to \cite{biswas2014episodic}. However, they computed their short-term and long-term features with ray-tracing in a 2D map, while we predict them directly in the frame based on the appearance of the points.

ICP-based SLAM algorithms are widely used in robotics, with many variants \cite{pomerleau2013comparing, zhang2014loam, mendes2016icp, deschaud2018imls}. We designed our PointMap SLAM with a focus on simplicity and efficiency. Similarly to \cite{deschaud2018imls}, we keep a point cloud with normals as the map, but we update the normals directly on the lidar frames with spherical coordinate neighbourhoods. We do not keep a pose graph and only rely on the frame-to-map alignment for localization. We do not need to take loop closure into account, given the relatively small size of the environment and the high precision of the method.

Computing occupancy probabilities with ray-casting has also been a common technique in the literature. Used at first for 2D occupancy grid mapping \cite{moravec1985high}, it was later adapted for 3D mapping \cite{izadi2011kinectfusion, hornung2013octomap}. In our case, PointRay computes occupancy probabilities on a point cloud instead of a grid, similarly to \cite{pomerleau2014long}, and therefore only models free space where points have been measured. Our main addition is the notion of long-term and short-term movables, which we get by combining multiple sessions. Other minor differences include a simplification of the probability update rules and the use of frustums instead of cones around lidar rays.

To the best of our knowledge, our approach is the first to use multi-session SLAM and ray-tracing as annotation tools for the training of semantic segmentation networks.


\section{Approach}
\label{sec_Method}

This section describes the key components of our self-supervised learning approach. First, we present the two main modules, PointMap and PointRay. Then we explain how the lidar frames of each successive session are annotated. Finally, we detail the network architecture used for predictions and its training.

\subsection{PointMap: ICP-based 3D Localization and Mapping}

We design an ICP-based SLAM algorithm to provide a point cloud map that can be used for the annotation process. ICP-based mapping approaches have been very successful, and are particularly well-suited for indoor environments, which contain a lot of planar surfaces with walls and man-made objects. Our SLAM algorithm, PointMap, has two components: an ICP solution, which aligns a frame on a map, and a mapping function that updates the map with the aligned frame. 

For a detailed description of the ICP algorithm, we refer to the in-depth review of Pomerleau et al. \cite{pomerleau2013comparing}. Following their work, we use the same elements to characterize our ICP: the data filters, the matcher, the outlier rejection, the distance function, and the convergence tests. Our choices for each element are listed in Table \ref{Table_icp}. Most of these are based on previous works \cite{mendes2016icp, deschaud2018imls}, taking into account that we use a Velodyne HDL-32E sensor. Some elements including matching, are simplified for efficiency. We use the latest odometry of the robot as the initial pose to solve the initialization issue common to most ICP solutions.

The second component of PointMap, the map update function, adds the information from an aligned frame to the map. To regulate the density of points, the map point cloud is paired with a sparse 3D grid, implemented with a hashmap. The grid is defined by an origin $x_{\mathrm{origin}} = (0, 0, 0) \in \mathbb{R}^{3}$ and a cell size ${dl}_{\mathrm{map}} = 3$cm. Only one point can be kept per voxel, with its normal and a score $s \in [0, 2] \subset \mathbb{R}$. The update rules are very simple. We keep the first point ever recorded in each voxel instead of a barycenter, to reduce the drift when adding more and more frames to the map. For a point $x \in \mathbb{R}^{3}$, the corresponding voxel is

\begin{equation}
\label{eq1}
    \mathrm{vox}(x) = \left\lfloor \frac{x - x_\mathrm{origin}}{{dl}_{\mathrm{map}}} \right\rfloor \in \mathbb{Z}^{3}.
\end{equation}

\begin{table}[t]
\caption{ICP Configuration for PointMap with a Velodyne HDL-32E.}
\setlength\tabcolsep{0.5pt}
\begin{footnotesize}
\begin{center}
\begin{tabular}{ L{1.6cm}  L{6.8cm}   }
\Xhline{2\arrayrulewidth}
\textbf{Elements} & \textbf{Choices} \TBstrut\\
\Xhline{2\arrayrulewidth}
Filters & Subsample frame with a 10cm grid. \TBstrut\\
\hline
Matcher & Only match 600 random points at each iteration.\Tstrut\\
 & Single nearest neighbour. \TBstrut\\
\hline
Rejection & Remove pairs if point-to-point distance is greater than 2m. \Tstrut\\
 & Remove pairs if point-to-plane distance is greater than 30cm (except for the first iteration). \Bstrut\\
\hline
Distance & Optimize point-to-plane distance. \TBstrut\\
\hline
Convergence & Stop at a maximum of 100 iterations. \Tstrut\\
 & Stop if relative motion goes below 0.01m and 0.001rad. \Bstrut\\
\Xhline{2\arrayrulewidth}
\end{tabular}
\end{center}
\end{footnotesize}
\label{Table_icp} 
\vspace{-3ex}
\end{table}

\noindent However, normals are constantly updated: we choose the normal with the highest score. We compute the frame normals using radius neighbours in spherical coordinates. A radius of $1.5 \times \theta_\mathrm{res}$ (the sensor angular resolution) is chosen so that each neighbourhood includes the points from three consecutive lines of a scan. We orient normals by choosing the incidence angle, $\alpha$, to be acute. Eventually a heuristic score translates two ideas: normals should ideally be computed from a close distance, $R$, but only if the lidar position is `facing' the surface:

\begin{equation}
    s = \left\{ 
    \begin{array}{ll}
        \left(\frac{\pi}{2} - \theta\right) / \left(\frac{\pi}{2} - \theta_0\right) & \mbox{if } \theta > \theta_0 \\
        1 + \exp{-(R - R_0)^2}  & \mbox{else}
    \end{array}
    \right. ,
\end{equation}

\noindent with $\alpha_0 = \frac{5\pi}{12}$, and $R_0 = 2$m. $\alpha_0$ is the minimum incidence angle for the lidar to be considered facing the surface, and $R_0$ the ideal distance to estimate the normal if the lidar is facing the surface.

For efficiency, we update the map only with the subsampled frame points used in ICP, therefore reducing the number of normals to be computed. We do not need all the points as the map will grow over time with multiple frames. This PointMap SLAM performs very well for our application, especially when using a previously mapped point cloud of the environment. We favour using an initial map in most of our experiments to ensure a good alignment between the different sessions.

\subsection{PointRay: Ray-tracing Movable Probabilities}

Movable probability is the opposite of occupancy probability. If a location exists in the map but is not occupied during the whole session it means it belongs to a movable object. We can retrieve movable probabilities in a map thanks to the data provided by lidar frames. Indeed, each frame provides two kinds of information: occupied space where the points are located, and free space along the lidar rays. We follow the idea from \cite{pomerleau2014long} and use the projection of the map in the frame spherical coordinates to model the lidar rays. The movable probabilities are deduced from the distance gap between frame points and map points. 

To each point of the map, $x_i$, our method, PointRay, assigns a set of $n_i$ binary values, $p^k_{i}$, $n_i$ being the number of times this point has been seen, by processing a sequence of lidar frames. For one frame, we first get the list of occupied voxels with Eq. \ref{eq1}. If $n_i$ has already been incremented for this voxel $i$, we add a new value $p^{n_i}_{i} = 0$ and increment $n_i$ by 1. We also update its direct neighbours in the grid, in the same way, to help with the lidar frame sparsity. Then, we project the frame into a frustum grid to encode free space. The frustum grid is a 2D grid in the $\theta$ and $\phi$ spherical dimensions, whose pixels store the smallest point distances to the lidar origin ($\rho$ spherical coordinate). It can be seen as the lidar depth image, whose resolution is defined by the parameters $d\theta = 1.33^\circ$ and $d\phi = 0.1^\circ$ (HDL-32E angular resolution). The map is also projected in the same frustum grid, after being cropped to the frame dimensions to avoid unnecessary computations and aligned in the frame pose. At this point, we can compare the radius $\rho$ of each point of the cropped map, to the depth saved in its corresponding frustum $\rho_0$. To be updated as free space, the point has to meet two conditions:

\begin{equation}
\label{eq2}
    \left.
    \begin{array}{ll}
    \mathrm{cond}_A: &\rho < \rho_0 - \mathrm{margin}(\rho_0) \\
    \mathrm{cond}_B: & \left|n_z\right| > \cos(\beta_{\mathrm{min}}) \quad \mathrm{OR} \quad \alpha < \alpha_{\mathrm{max}}
    \end{array}
    \right. ,
\end{equation}

\noindent where $\mathrm{margin}(\rho_0) = \rho_0 \max(d\theta, d\phi) / 2$ is the largest half size of the frustum at this particular range, $n_z$ is the vertical component of the point normal, and $\alpha$ the incidence angle of the lidar ray with this normal. $\alpha_{\mathrm{max}} = \frac{5\pi}{12}$ and $\beta_{\mathrm{min}} = \frac{\pi}{3}$ are heuristic thresholds. $\mathrm{cond}_A$ ensures that, at any range, a planar surface whose incidence angle is less than 45$^{\circ}$, will not be updated as free space. $\mathrm{cond}_B$ handles the wider incidence angles, we do not want to update for extreme incidence values, except if the normal is vertical because tables are usually nearly parallel to the lidar rays and would never be updated otherwise. Because of $\mathrm{cond}_B$, ground points are more likely to have high movable probabilities, but we take care of that by extracting the ground as a distinct semantic class, unaffected by ray-tracing. If $\mathrm{cond}_A$ and $\mathrm{cond}_B$ are met, and only if it has not been updated yet, the map point is assigned a new value $p^{n_i}_{i} = 1$ and $n_i$ is incremented by 1.

When a full session has been ray-traced, we can get the resulting movable probabilities for every point $x_i$ of the map:

\begin{equation}
    p_\mathrm{mov}(x_i) = \left\{ 
    \begin{array}{ll}
        \sum\limits_{k<n_i}{p^k_{i}} / n_i & \mbox{if } n_i > n_\mathrm{min} \\
        0.5  & \mbox{else}
    \end{array}
    \right.  .
\end{equation}

\noindent We do not consider a point movable if it has not been seen by at least $n_\mathrm{min} = 10$ times. We now have all the tools we need to design our multi-session annotation process.

\begin{figure*}[t]
    \centering
    \includegraphics[width=0.999\textwidth, keepaspectratio=true]{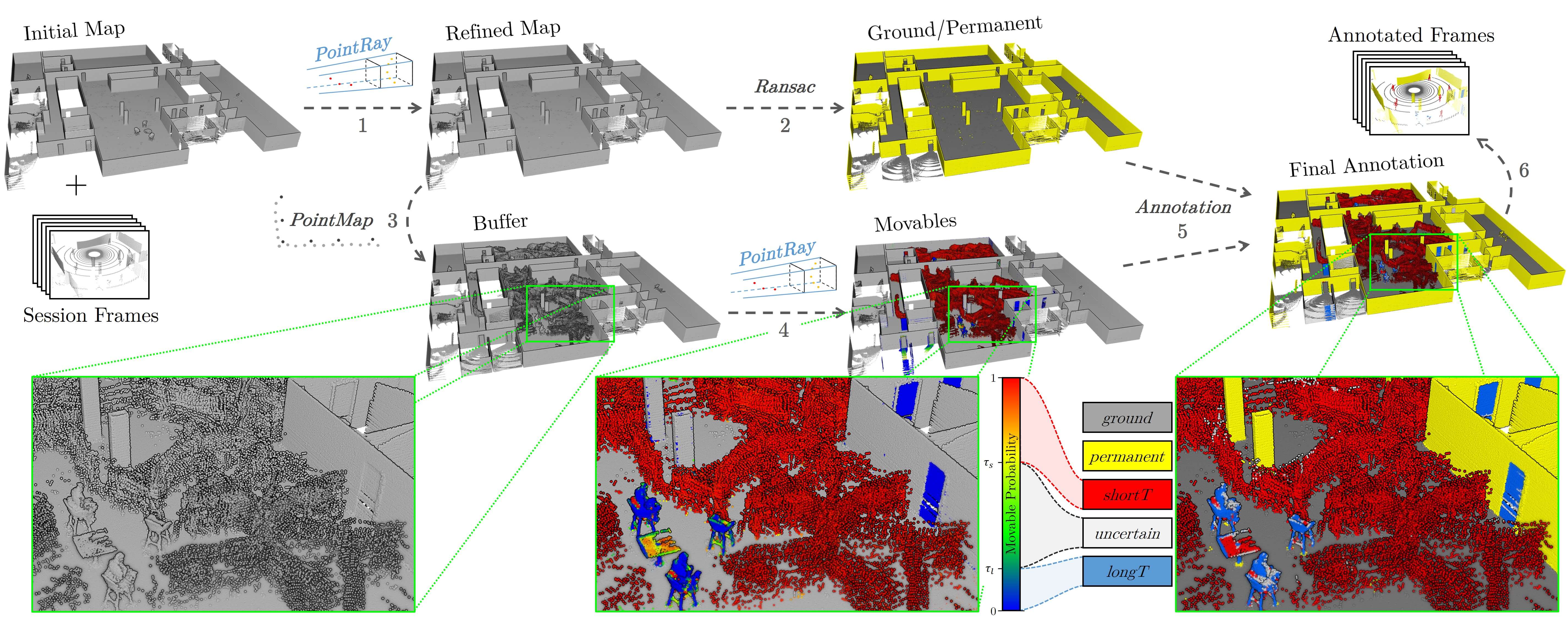}
    \caption{Illustration of our annotation process for the frames of one session. With PointRay, movables are erased from the initial map. We get ground and permanent points with RANSAC, and the buffer cloud is mapped. Then, points that are neither ground nor permanent are annotated given their movable probability, obtained with PointRay. Finally, the buffer annotations are projected back to the frames.}
    \label{fig_annot}
    \vspace{-3ex}
\end{figure*}

\subsection{Multi-session Ray-tracing Annotation}

With PointMap and PointRay we can compute movable probabilities on the map of one session. In this section, we explain how we extend this to annotate multiple sessions.

The first thing to consider when dealing with multiple sessions is the concept of movable objects. In a single session, if an object moves, its points in one frame can then be seen as free space by the other frames. Therefore, it is movable. However, imagine now that we perform a `cross-session' ray tracing, by checking what the frames from another session see. An object that was not moving, such as a chair, could have a different location the next day and thus be seen as movable. This is why we introduce the notion of short-term movables and long-term movables. Along with these two semantic classes, we classify the permanent points, which are the ones that remain non-movable across all sessions. We also add the ground label, as the ground is easy to extract in our indoor scenario.

Another issue to take into account is the session alignment. PointRay requires the frames to be perfectly aligned with the map we want to annotate. A simple way to ensure the alignment of multiple sessions is to have them localize against the same initial map of the environment. In a multi-session scenario, it is perfectly acceptable to suppose that the environment was mapped at the beginning of the project. In the following, we thus assume that an initial mapping tour has been conducted covering the whole experimental space.

We use this initial map as the starting point of our algorithm. It will be the recipient of the ground and permanent points. The process leading to the annotation of one session is described in Figure \ref{fig_annot}. We start by applying PointRay to the initial map (step 1), and by removing the points with a higher movable probability than $\tau_{\mathrm{refine}} = 0.9$. This refined map replaces the initial map at this point, and for all future sessions. As the map will be refined by all the sessions, we can afford to use a high threshold of $\tau_{\mathrm{refine}}$ and miss some movables. They will eventually be caught by another session. It is safer to avoid removing permanent points that should remain in the map. The points from the refined map are classified as \textit{permanent} or \textit{ground} (step 2) based on a ground extraction algorithm using RANSAC \cite{fischler1981random}.

In the next step, we use PointMap to create a map aggregating the points from all the frames of this session (step 3). This is a temporary map and will be discarded at the end of the process, hence its name: the buffer. The role of the buffer is to be annotated and then to transfer the annotations back to the frames via nearest-neighbour interpolation. As shown in the close-ups, PointRay is first used to get the movable probabilities of the buffer points that are not from the refined map (step 4). These probabilities are mapped to the two remaining classes, \textit{shortT} for $p_\mathrm{mov} > \tau_\mathrm{short}$, \textit{longT} for $p_\mathrm{mov} < \tau_\mathrm{long}$ (step 5). The rest of the points are annotated as \textit{uncertain}. In the end, we project the annotations back on the frames (step 6) to get a training set for our deep neural network.

\subsection{Network Training and Inference}
\label{subsec_network}

Our work aims to propose a strategy for the self-supervision of any lidar segmentation architecture. We chose the recent 3D point cloud convolutional network KPConv \cite{thomas2019kpconv}, as it has shown great performance on a large variety of tasks and offers a comprehensive source code in PyTorch. For the most part, we use the same parameters as the original KPConv architecture. For the rest, we explain the choices we made for the crucial parameters in the following.

KPConv was originally designed for full-scene segmentation instead of lidar-frame segmentation, and it uses a sphere sampling strategy at training: the network is trained on spherical subsets of the scene picked randomly. Although it may seem inappropriate to not use the full lidar frame at training, we found that this spherical input sampling works well in our case. It can be explained by the fact that it creates a kind of class balancing. Indeed, the picking strategy ensures that the same number of spheres is picked around all the classes of the dataset. Therefore, minority classes are seen more often and better classified.

We thus have two parameters to set regarding the input, the sphere radius $R_\mathrm{in}$, and the input subsampling grid size $dl_\mathrm{in}$. We chose the values $R_\mathrm{in}=4$m and $dl_\mathrm{in}=0.04$ to cover a non-negligible part of the lidar frame and to keep enough detail for the prediction to be accurate. Using a smaller part of the frame also allows us to have a batch size of $10$ input point clouds. Following the standard input augmentation strategies, we add random horizontal flipping and random rotation along the vertical axis to the input point cloud. For more details on the architecture and the training parameters, we refer to the original KPConv article \cite{thomas2019kpconv} and its supplementary material.

During inference, we feed the entire lidar frame to the network. The frame is first subsampled and the network predicts our four labels for each subsampled point. We then use a triaging system on this classified subsampled frame. Short-term movable points likely are from moving humans and should avoid the robot on their own. There is no point in planning around them, so we remove them from the global and local planning systems. For localization, we also remove long-term movables as the map does not contain them.

Note that it is not an issue that the network was trained on subsets of lidar frames, because, as it is carefully explained in the KPConv paper, the effective receptive field in the latest stage of the network does not exceed a few meters. We also want to highlight that the network is not optimized yet for real-time and can only process 2 frames per second in our setup. This is not a problem as we work in simulation for the time being. However, we are well aware of the bottlenecks in the implementation and have optimization in mind for real-world experiments in future works. In addition to optimizing the code, a trade-off between performance and computing speed can be set with the parameter $dl_\mathrm{in}$. It is possible to use a larger subsampling size to reach real-time, at the cost of loss in prediction accuracy.


\section{Experiments}
\label{sec_Exp}

\subsection{Experimental Setup}

\begin{figure}[t]
    \centering
    \includegraphics[width=0.99\columnwidth, keepaspectratio=true]{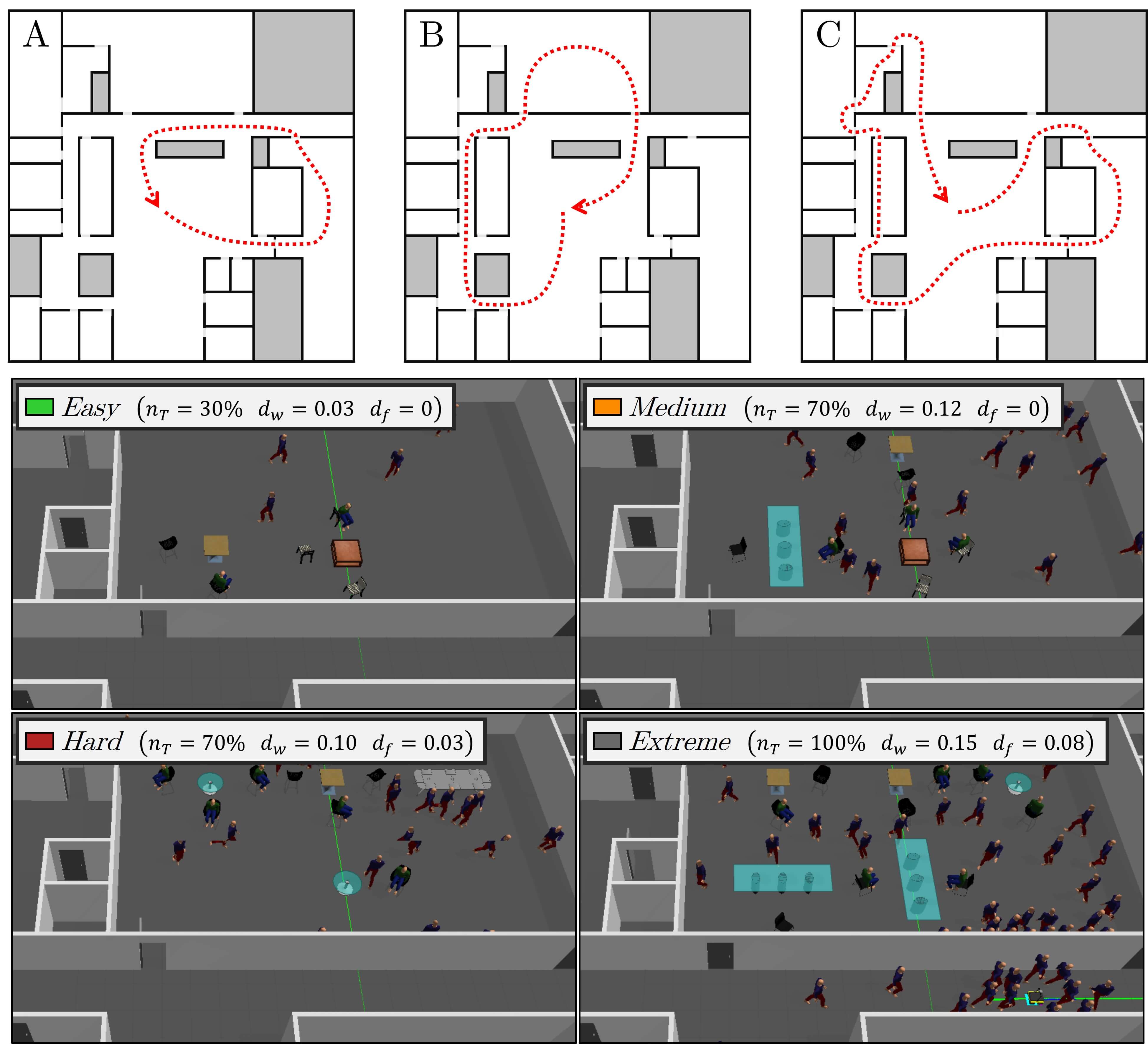}
    \caption{Visualization of our three tours (top) and our four scenarios in the Gazebo environment (bottom).}
    \label{fig_scenarios}
    \vspace{-3ex}
\end{figure}

We evaluate our self-supervised learning approach in a Gazebo simulation. We reproduced the original experimental space in the simulator. For the furniture, we chose a few models of chairs and tables from Gazebo's collection, that we can add to the world at preset positions. We can also populate the world with different types of actors, including sitting people, randomly walking people (wanderers), and people following a random point around the robot (followers). For our experiments, we designed increasingly difficult scenarios by changing the world parameters: The percentage of preset positions filled with tables and chairs $n_T$ and the wanders' and followers' population density, respectively $d_w$ and $d_f$, in people/m$^2$. Each scenario can be seen with its parameters in Figure \ref{fig_scenarios}. In addition, we also show three predefined tours, which consist of a succession of targets $t_k \in \mathbb{R}^2$ that the robot has to reach. Therefore, there are 12 possible sessions with different setups.

The experiments are conducted on a DGX station with an Intel(R)
Xeon(R) E5-2698 v4 CPU and four Tesla V100 DGXS GPU. The annotation process only uses CPU and computes a whole session of 1,000 lidar frames in less than 30 minutes on average, which is completely reasonable for an offline task. For the network training, we use a single GPU and reach convergence in about 10 hours. During the online sessions, we use PointMap as our localization algorithm. Its ROS implementation runs at more than 30 fps on average. However, as explained in paragraph \ref{subsec_network}, the network inference is much slower and takes around 500ms to process one frame. For now, we get around this issue by pausing the simulation when needed and are confident about optimization for future works.

\subsection{Self-Improvement Over Multiple Sessions}

\begin{figure}[b]
    \vspace{-2ex}
    \centering
    \includegraphics[width=0.98\columnwidth, keepaspectratio=true]{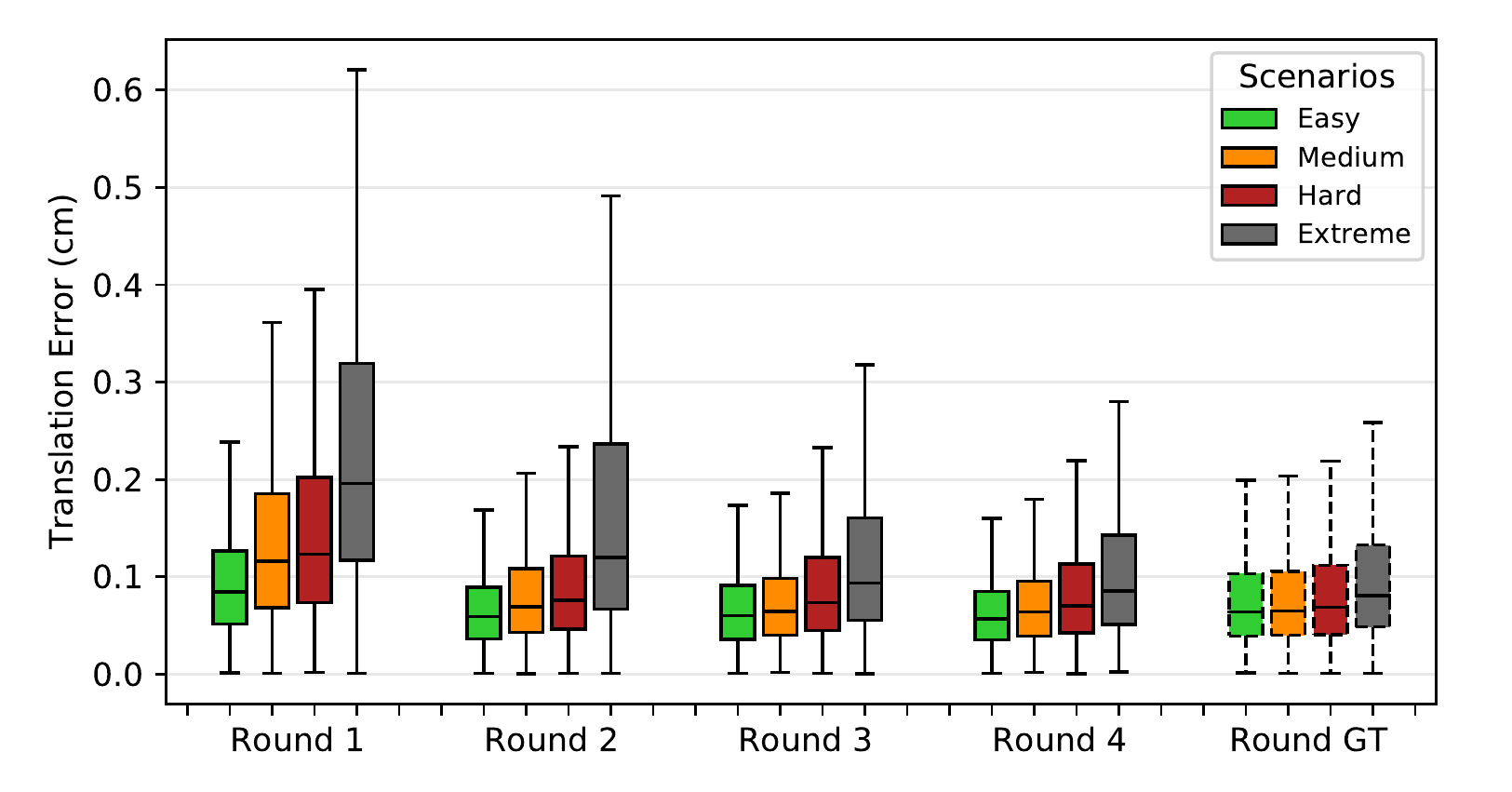}
    \caption{PointMap localization error for all rounds and scenarios, including an additional round using groundtruth point cloud filtering. Each box includes the three tours combined.}
    \label{fig_selfimprove}
\end{figure}

In our first experiment, we demonstrate the ability of our approach to improve itself over multiple sessions. We define precise experimental conditions under which the performance improvement can be evaluated. The robot has to navigate through the 12 possible sessions several times. Between each of these rounds, we evaluate the robot success rate, perform the annotation of the successful tours, and train the network to be used in the next round. The results of this experiment are shown in Table \ref{Table_success} and Figure \ref{fig_selfimprove}. 

\begin{table}[t]
\caption{Success rate of each round and each scenario averaged on tours A, B, and C. Underlined if all tours were completed.}
\setlength\tabcolsep{0.5pt}
\begin{footnotesize}
\begin{center}
\begin{tabular}{| C{1.5cm} | *{4}{C{1.2cm}} | C{1.2cm}  | }

\cline{2-6}
\multicolumn{1}{c|}{ } & Easy & Medium & Hard & Extreme & Avg\TBstrut\\

\hline
Round 1	& $\underline{100\%}$	& $81\%$	& $7\%$	& $3\%$	& $\mathbf{48\%}$	\TBstrut\\
Round 2	& $\underline{100\%}$	& $\underline{96\%}$	& $\underline{100\%}$	& $79\%$	& $\mathbf{94\%}$	\TBstrut\\
Round 3	& $\underline{100\%}$	& $\underline{100\%}$	& $\underline{100\%}$	& $\underline{100\%}$	& $\mathbf{100\%}$	\TBstrut\\
Round 4	& $\underline{100\%}$	& $\underline{100\%}$	& $\underline{100\%}$	& $\underline{100\%}$	& $\mathbf{100\%}$	\TBstrut\\

\hline
\end{tabular}
\end{center}
\end{footnotesize}
\vspace{-3ex}
\label{Table_success}
\end{table}

We define the success rate of a tour as the percentage of targets $t_k$ reached by the robot. Sometimes, the robot can miss a target but still complete the rest of the tour. The success rate of each round can be seen in Table \ref{Table_success}. We find that the robot is only able to complete all the tours in the \textit{easy} scenario at the beginning. We thus annotate and train the network only on the tours of this scenario. With this first network, we see a large improvement in round 2, as the robot completes all the tours in the \textit{easy}, \textit{medium} and \textit{hard} scenarios. Then, with the second network trained on these three scenarios, the robot is eventually able to complete the \textit{extreme} scenario. A final round is performed with the third network trained on all the tours and scenarios from round 3. Examples of classified frames are shown in Figure \ref{fig_frames}.

In addition to the success rate we see a noticeable improvement of the localization error for each new round, especially in the \textit{extreme} scenario. In this controlled simulated environment, the robot reaches excellent performance in a few rounds. The more complex the scenario, the more rounds it needs. We can assume that our approach would never stop improving itself in an uncontrolled real scenario, where unexpected situations can happen at any time.

\subsection{Network Predictions Evaluation}

\begin{figure}[t]
    \centering
    \includegraphics[width=0.99\columnwidth, keepaspectratio=true]{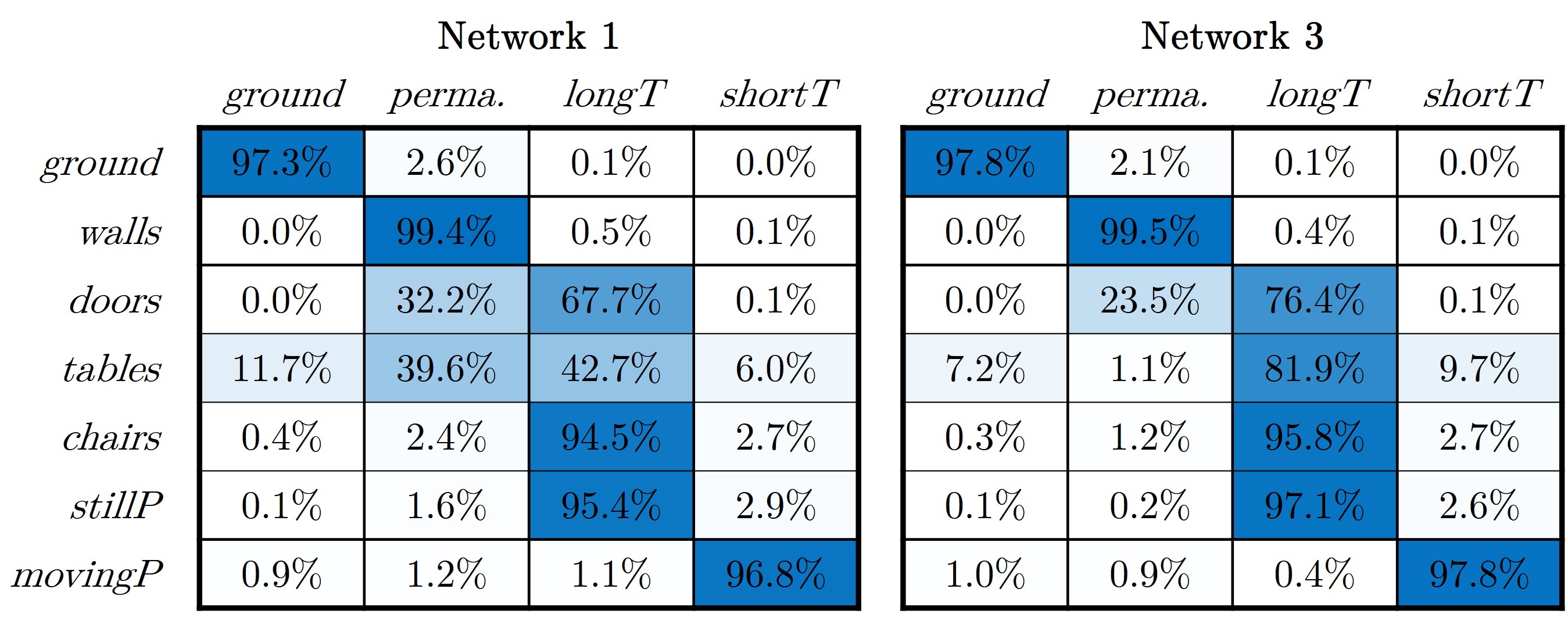}
    \caption{Confusion matrices between groundtruth labels and our networks predictions, normalized by rows.}
    \label{fig_confusion}
    \vspace{-2ex}
\end{figure}

\begin{figure}[b]
    \centering
    \vspace{-2ex}
    \includegraphics[width=0.99\columnwidth, keepaspectratio=true]{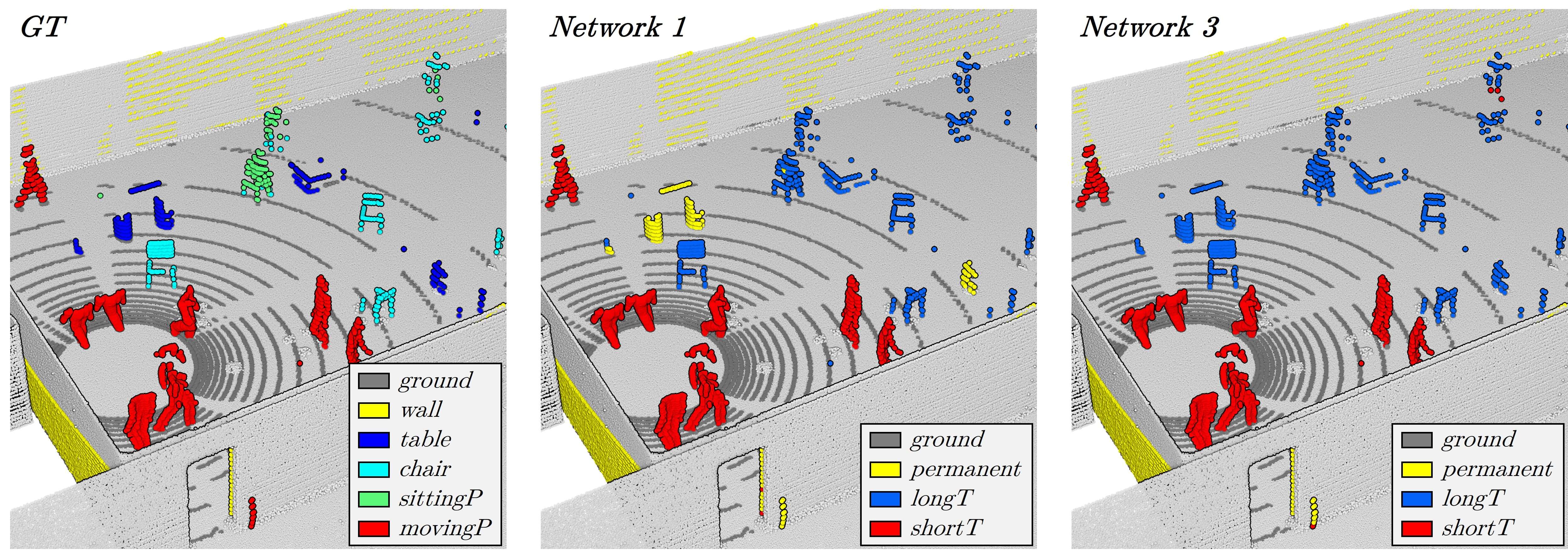}
    \caption{Example of classified frames from the networks trained on round 1 and round 3, compared to groundtruth labels. Map points are shown in light grey for visualization purposes.}
    \label{fig_frames}
\end{figure}

Using the simulator to provide groundtruth labels of the points, we evaluate our network predictions in this second experiment. The groundtruth characterizes the types of objects in the scene, which is different from our annotation, but useful to understand what happens between the rounds of the first experiment. As shown in Figure \ref{fig_confusion}, the scenes contain \textit{ground}, \textit{walls}, \textit{doors}, \textit{tables}, \textit{chairs}, \textit{still people} and \textit{moving people}. We compare these groundtruth classes to the predictions of our first and third networks in the form of confusion matrices. As expected, most of the points from \textit{ground}, \textit{wall}, and \textit{moving people} are predicted as \textit{ground}, \textit{permanent}, and \textit{short-term}, respectively, by our networks, with a recall of at least $95\%$. The rest of the classes are predicted in the majority as \textit{long-term}, but not as confidently. For example, the first network seems to struggle with tables, sometimes identifying them as \textit{permanent} points. We can explain this by the fact that the network has not `seen' every type of table that our scenarios possibly have, and thus confuses them with walls. This particular behavior can be seen in Figure \ref{fig_frames}, where two of the tables are predicted as permanent points by the first network. This example shows how our method adapts when seeing new objects or situations.

Finally, we evaluate the localization error of our robot when using these groundtruth labels for an additional round. As shown in Figure \ref{fig_selfimprove}, compared to our best network, using the groundtruth labels gives similar results. It confirms that our method converges in a few rounds.

\subsection{Improvement on Other Localization Methods}

\begin{figure}[t]
    \vspace{-2ex}
    \centering
    \includegraphics[width=0.99\columnwidth, keepaspectratio=true]{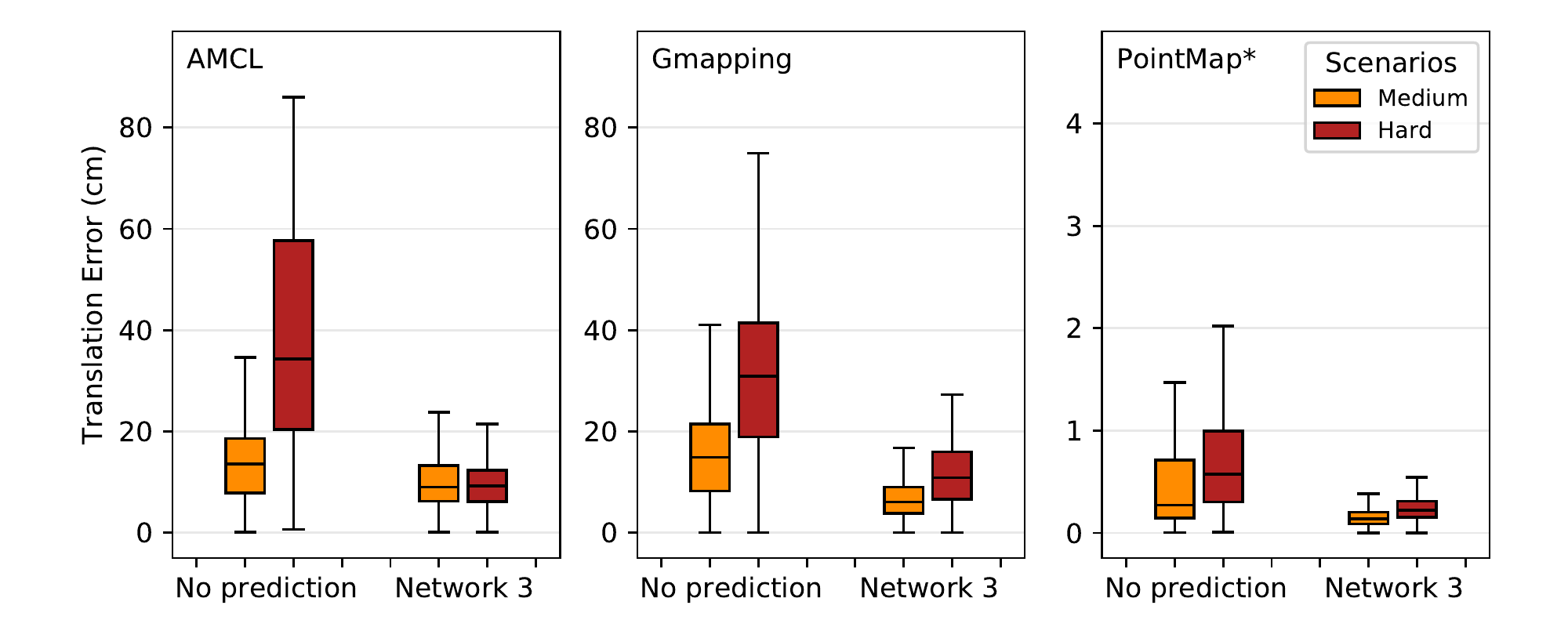}
    \caption{Localization error for AMCL, Gmapping and PointMap$^*$, with and without our network predictions.}
    \label{fig_amcl}
\end{figure}

Our last experiment shows that our approach can generalize to other common navigation systems. Our benchmark includes a localization method, AMCL \cite{fox2003adapting}, and two SLAM algorithms, Gmapping \cite{grisetti2007improved}, and PointMap$^*$ (PointMap without an initial map). We compare the localization performances of these methods when using our network predictions to filter point clouds or not. We conduct this experiment in the \textit{medium} and \textit{hard} scenarios and with our best performing network. In Figure \ref{fig_amcl}, we notice that our network predictions improve the performance of every localization method we tested. The margin of improvement for the \textit{hard} scenario is the biggest with a localization error divided by $3.8$, $3.0$, and $2.7$ for AMCL, Gmapping, and PointMap$^*$, respectively.

\section{Conclusion} 
\label{sec_Conclusion}

In this paper, we presented the first self-supervised learning approach for the semantic segmentation of lidar frames. We showed that it is possible to learn semantic classes without human annotation and use them to improve navigation systems. Experiments conducted in the simulation were valuable for the insight offered by having a groundtruth available, but we also want to carry out experiments in the real world. We gave a first idea on how to use these semantic labels and showed how efficient it can be. We hope to pave the way for future work investigating how self-supervised learning can help more elaborate navigation tasks with object tracking or goal prediction, for example.

\addtolength{\textheight}{-9.5cm}   


\newpage

\bibliographystyle{IEEEtran}
\bibliography{ICRA2018}

\end{document}